\documentclass[conference]{IEEEtran}
\IEEEoverridecommandlockouts

\ifCLASSINFOpdf
\else
\fi

\usepackage{url}
\usepackage{algorithm}
\usepackage{algorithmic}
\usepackage[dvips]{graphicx}
\usepackage{subfigure}
\usepackage{amsmath,amssymb,amsfonts}
\usepackage{fancyhdr}

\newcommand{\bvec}[1]{\mbox{\boldmath $#1$}}

\begin{document}
\title{Automatic Extraction of Road Networks from Satellite Images by using \\Adaptive Structural Deep Belief Network
\thanks{\copyright 2021 IEEE. Personal use of this material is permitted. Permission from IEEE must be obtained for all other uses, in any current or future media, including reprinting/republishing this material for advertising or promotional purposes, creating new collective works, for resale or redistribution to servers or lists, or reuse of any copyrighted component of this work in other works.}
}

\author{
\IEEEauthorblockN{Shin Kamada}
\IEEEauthorblockA{Research Organization of Regional Oriented Studies,\\
Prefectural University of Hiroshima\\
1-1-71, Ujina-Higashi, Minami-ku, \\
Hiroshima 734-8558, Japan\\
E-mail: skamada@pu-hiroshima.ac.jp}
\and
\IEEEauthorblockN{Takumi Ichimura}
\IEEEauthorblockA{Faculty of Management and Information System,\\
Prefectural University of Hiroshima\\
1-1-71, Ujina-Higashi, Minami-ku, \\
Hiroshima 734-8558, Japan\\
E-mail: ichimura@pu-hiroshima.ac.jp}
}

\maketitle

\thispagestyle{plain}




\pagestyle{fancy}{
\fancyhf{}
\fancyfoot[R]{}}
\renewcommand{\headrulewidth}{0pt}
\renewcommand{\footrulewidth}{0pt}

\begin{abstract}
In our research, an adaptive structural learning method of Restricted Boltzmann Machine (RBM) and Deep Belief Network (DBN) has been developed as one of prominent deep learning models. The neuron generation-annihilation in RBM and layer generation algorithms in DBN make an optimal network structure for given input during the learning. In this paper, our model is applied to an automatic recognition method of road network system, called RoadTracer. RoadTracer can generate a road map on the ground surface from aerial photograph data. In the iterative search algorithm, a CNN is trained to find network graph connectivities between roads with high detection capability. However, the system takes a long calculation time for not only the training phase but also the inference phase, then it may not realize high accuracy. In order to improve the accuracy and the calculation time, our Adaptive DBN was implemented on the RoadTracer instead of the CNN. The performance of our developed model was evaluated on a satellite image in the suburban area, Japan. Our Adaptive DBN had an advantage of not only the detection accuracy but also the inference time compared with the conventional CNN in the experiment results.
\end{abstract}



\IEEEpeerreviewmaketitle

\section{Introduction}
In recent years, Artificial Intelligence (AI) has shown remarkable development centering on deep learning \cite{Bengio09}. With major achievements of recent deep learning techniques, they learn to realize a part of human cognitive ability for image recognition, speech recognition, and so on. Moreover, large amounts of data can be collected to make decision and prediction by using the applied deep learning models in the real world problems.

The adaptive structural learning method of Deep Belief Network (Adaptive DBN) \cite{Kamada18_Springer} has been developed, which has an outstanding function to find the optimal network structure for given input signals. The proposed method can firstly determine an optimal number of neurons for an Restricted Boltzmann Machine (RBM) \cite{Hinton06, Hinton12} by the neuron generation-annihilation algorithm \cite{Kamada16_SMC, Kamada16_ICONIP}. Moreover, an optimal number of layers of DBN can be also generated during the training \cite{Kamada16_TENCON}. Adaptive DBN method shows the highest classification capability for image recognition among some benchmark datasets such as CIFAR-10, CIFAR-100 \cite{CIFAR10}, and so on. The paper \cite{Kamada18_Springer} reported that the model can reach higher classification accuracy for test cases than the traditional Convolutional Neural Networks (CNNs) such as AlexNet \cite{AlexNet}, VGG16 \cite{VGG16}, GoogLeNet \cite{GoogLeNet}, and ResNet \cite{ResNet}.

In this paper, our developed Adaptive DBN is applied to an automatic recognition method of road map on the ground surface from aerial or satellite image. Building a road map and its maintenance require a lot of cost by human experts due to the complicated tasks. For the problem, several detection systems to the road map using deep learning was developed since 2017 \cite{Liu18, He20, Lian20, Tan20}. DeepRoadMapper \cite{Mattyus17} is a segmentation technique using deep learning to automatically detect roads from a satellite image. However, it was not able to achieve high detection accuracy, since some non-road features such as trees, river, and shadow of buildings covered roads and they became noise. In the other words, the model has insufficient segmentation power to clearly distinguish the road from the non-road features from only aerial image. In order to solve the problem, the automatic recognition method of road network using deep learning called RoadTracer \cite{Bastani18} was proposed. The iterative graph search algorithm is used to find network graph representing the connectivity between roads, which is a different idea of segmentation based method.

However, the iterative graph search takes a long calculation time for not only the training phase but also the inference phase, because its search algorithm requires iterative recognition results by using a CNN for the images. If the CNN does not reach high accuracy, the search algorithm cannot find the next road and the algorithm will stop soon during the search process. In order to improve the image recognition power, Adaptive DBN was applied into the RoadTracer instead of the CNN. In this paper, the road detection for the suburban area including many trees or complicated small roads was challenged. The Adaptive DBN had an advantage of not only the detection accuracy but also the inference time compared with the conventional CNN with the different search parameter.

The remainder of this paper is organized as follows. In the section \ref{sec:adaptive_dbn}, the basic idea of Adaptive DBN is briefly explained. In the section \ref{sec:roadtracer}, the basic behavior of searching algorithm in RoadTracer is described. In the section \ref{sec:roadtracer_dbn}, our Adaptive DBN is implemented on the RoadTracer and the effectiveness of our proposed model is verified on a test satellite image. In the section \ref{sec:conclusion}, some discussions are given to conclude this paper.

\section{Adaptive Structural Learning Method of Deep Belief Network}
\label{sec:adaptive_dbn}

\subsection{RBM and DBN}
\label{sec:RBMDBN}
\begin{figure}[bt]
\centering
\includegraphics[scale=0.8]{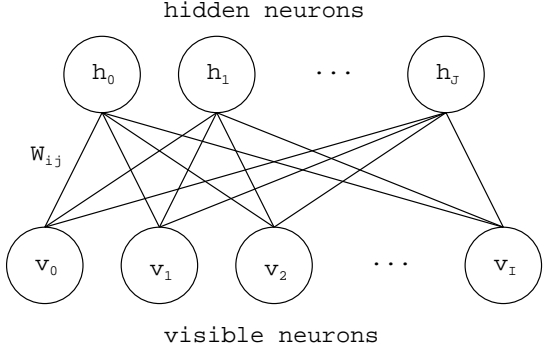}
\caption{Network structure of RBM}
\label{fig:rbm}
\end{figure}

An RBM \cite{Hinton12} is an unsupervised learning to learn a probability distribution of input using two kinds of binary layers as shown in Fig.~\ref{fig:rbm}: a visible layer $\bvec{v} \in \{0, 1 \}^{I}$ for an input and a hidden layer $\bvec{h} \in \{0, 1 \}^{J}$ for a feature vector. The three kinds of learning parameters $\bvec{\theta}=\{\bvec{b}, \bvec{c}, \bvec{W} \}$ are trained for the given input to minimizes the energy function $E(\bvec{v}, \bvec{h})$ as follows.
\begin{equation}
E(\bvec{v}, \bvec{h}) = - \sum_{i} b_i v_i - \sum_j c_j h_j - \sum_{i} \sum_{j} v_i W_{ij} h_j ,
\label{eq:energy}
\end{equation}
\begin{equation}
p(\bvec{v}, \bvec{h})=\frac{1}{Z} \exp(-E(\bvec{v}, \bvec{h})),
\label{eq:prob}
\end{equation}
\begin{equation}
Z = \sum_{\bvec{v}} \sum_{\bvec{h}} \exp(-E(\bvec{v}, \bvec{h})),
\label{eq:z}
\end{equation}
where $b_{i}$ and $c_{j}$ are the biases for $v_{i}$ and $h_{j}$. $W_{ij}$ is the weights between them. Eq.~(\ref{eq:prob}) is a probability of $\exp(-E(\bvec{v}, \bvec{h}))$. $Z$ is calculated by summing energy for all possible pairs of visible and hidden vectors in Eq.~(\ref{eq:z}). A standard estimation for the parameters is maximum likelihood in a statistical model $p(\bvec{v})$.

A DBN \cite{Hinton06} forms a hierarchical network which is constructed by stacking several pre-trained RBMs. The activation values of hidden neurons at $l-1$-th RBM are taken to next level of input at $l$-th RBM as Eq.~(\ref{eq:prob_dbn}).
\begin{equation}
\label{eq:prob_dbn}
p(h_j^{l} = 1 | \bvec{h}^{l-1})= sigmoid(c^{l}_j + \sum_{i}W^{l}_{ij} h^{l-1}_{i}),
\end{equation}
where $c^{l}_j$ and $W^{l}_{ij}$ indicate the parameters for the $j$-th hidden neuron and the weight at the $l$-th RBM, respectively. $\bvec{h}^{0} = \bvec{v}$ means the given input data. After the construction of the pre-trained RBMs, an output layer can be appended to the last RBM layer for supervised learning, then the corresponding weights are fine-tuned.

\begin{figure}[tbp]
\begin{center}
\subfigure[Neuron generation]{\includegraphics[scale=0.5]{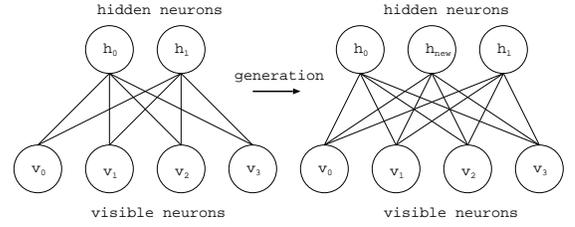}\label{fig:neuron_generation}}
\subfigure[Neuron annihilation]{\includegraphics[scale=0.5]{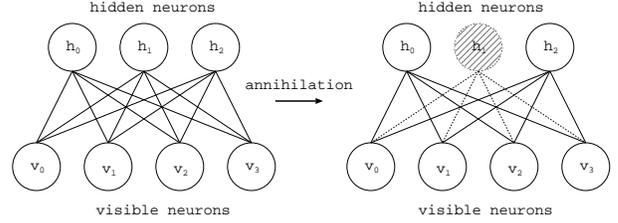}\label{fig:neuron_annihilation}}
\vspace{-3mm}
\caption{Adaptive RBM}
\label{fig:adaptive_rbm}
\end{center}
\end{figure}

\begin{figure*}[tbp]
\centering
\includegraphics[scale=0.8]{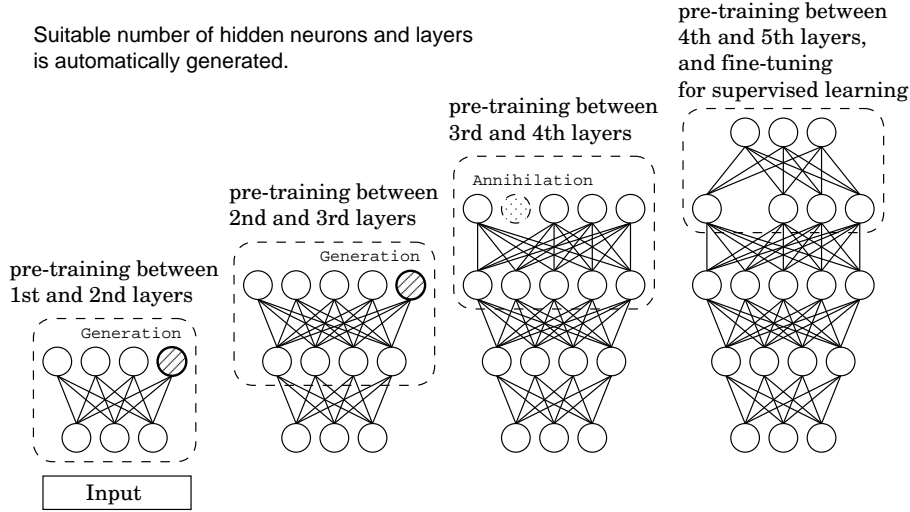}
\vspace{-3mm}
\caption{An Overview of Adaptive DBN}
\label{fig:adaptive_dbn}
\end{figure*}

\subsection{Neuron Generation and Annihilation Algorithm of RBM}
\label{subsec:adaptive_rbm}
The optimal design for the size of network architecture is important problem to achieve high performance in deep learning models. In our research, we have developed an adaptive structural learning method of RBM, called Adaptive RBM \cite{Kamada18_Springer}, to find the optimal number of hidden neurons for given input during the learning. The key idea of Adaptive RBM is the observation of Walking Distance (WD), which means a criterion how the model's parameters are changed on the iterative learning \cite{Kamada16_SMC}. If any fluctuation at a neuron in the network is observed in the training process, the corresponding neuron will be copied and generated as a new neuron, as shown in Fig.~\ref{fig:neuron_generation}.

Moreover, the neuron annihilation algorithm is developed to remove the redundant neurons after the certain number of neurons are generated. Fig.~\ref{fig:neuron_annihilation} shows that the corresponding neuron is annihilated \cite{Kamada16_ICONIP}. Due to a cross check system, please see the paper \cite{Kamada18_Springer} for detailed algorithm of the method.

\subsection{Layer Generation Algorithm of DBN}
\label{subsec:adaptive_dbn}

Based on the neuron generation algorithm, we have developed a hierarchical model of Adaptive DBN which can automatically generate a new Adaptive RBM in the training process as shown in Fig.~\ref{fig:adaptive_dbn}. Since DBN is a stacking box of pre-trained RBMs, the total RBMs' energy values and WD can be used as a criterion to the layer generation. If the criterion is larger than the pre-determined threshold, a new RBM layer will be generated to complement lack of representation for given input data \cite{Kamada18_Springer}.

\section{RoadTracer}
\label{sec:roadtracer}
\subsection{Graph Search Algorithm}
\label{subsec:roadtracer_alg}
RoadTracer \cite{Bastani18} is an automatic recognition method of a road map on the ground surface from aerial photograph data. The iterative graph search algorithm is used to find network graph connectivities between roads using graph structure, which is different idea of segmentation based method of deep learning. {\bf Algorithm \ref{alg:Ite_Graph_Construction}} shows a pseudo code of the search algorithm. The search algorithm firstly initializes any starting location of search $v_0$, an empty graph $G$, and a stack of vertices $S$. $G$ is used to store detected vertices and edges. $S$ is used to stack recent detected vertices as searching history. Secondly, for current searching position $S_{top}$, the decision function takes an action that is walk or stop, and then updates $G$ and $S$ according to the decided action. This searching step is repetitively conducted until the terminate condition is satisfied.

The detailed specification of the decision function is as follows. The input of the decision function is the current graph $G$, the current searching position $S_{top}$, and an aerial or satellite image around $S_{top}$. For the input, the function takes two kinds of outputs, which are action and angle. The action is defined as `walk' or `stop'. If `walk' is determined, the current searching position $S_{top}$ is moved to the decided angle in a certain distance, and the algorithm updates $G$ and $S$. If `stop' is determined, the current searching position $S_{top}$ is popped from the stack of vertices $S$ to back the previous position in next step.

\begin{figure*}[tbp]
  \centering
  \includegraphics[scale=0.8]{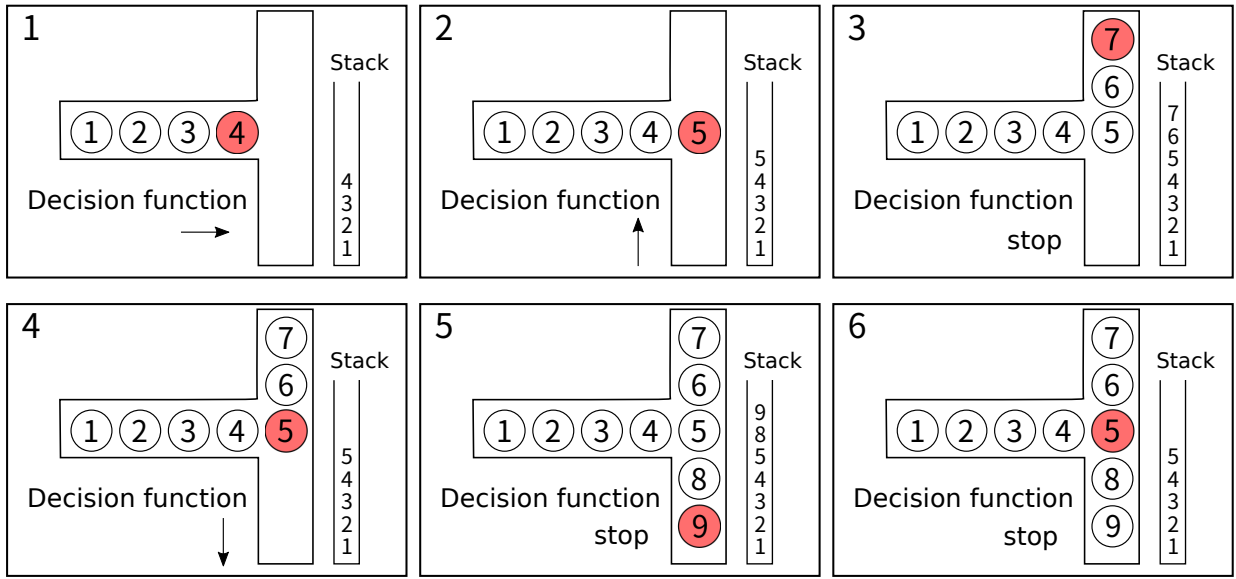}
  \vspace{-3mm}
  \caption{Example of Search Algorithm}
  \label{fig:SearchAlgorithm}
\vspace{-5mm}
\end{figure*}

Fig.~\ref{fig:SearchAlgorithm} is an example of searching behavior in an intersection. Firstly, the vertices 1 to 4 are stacked in $S$, and the decision function takes `walk' to east direction, then the vertex 5 is stacked in $S$. Secondly, the decision function takes `walk' to north direction as same procedure. Thirdly, the function takes `stop' because the available vertices are not found. After that, the current position is moved to back the intersection and the algorithm searches the south area. Finally, the stop action is determined again, then the search algorithm is finished.

\begin{algorithm}[tbp]
\caption{Graph Search \cite{Bastani18}}
\label{alg:Ite_Graph_Construction}
\begin{algorithmic}[1]
\REQUIRE A starting location $v_{0}$ and the bounding box $B$
initialize graph $G$ and vertex stack $S$ with $v_{0}$
  \WHILE {$S$ is not empty}
    \STATE \textit{action}, $\alpha$ := decision func($G$, $S_{top}$,\textit{Image})
    \STATE $u$ := $S_{top} + (D \cos\alpha, D\sin\alpha)$
    \IF {action = stop or $u$ is outside $B$}
    \STATE  pop $S_{top}$ from $S$
    \ELSE
    \STATE  add vertex $u$ to $G$
    \STATE  add an edge $(S_{top}, u)$ to $G$
    \STATE  push $u$ onto $S$
    \ENDIF
\ENDWHILE
\end{algorithmic} 
\end{algorithm}

\subsection{Decision function of CNN}
\label{subsec:roadtracer_cnn}
Deep learning is used in the inference process of the decision function in RoadTracer. The original paper \cite{Bastani18} constructed a CNN with 17 convolutional layers in addition to input and output layers. The input layer is $d \times d \times 4$ architecture which consists of $d \times d \times 3$ RGB image and graph $G$ around the current position. For the input, `walk' and `stop' in the action are represented by two neurons $O_{action} = (O_{walk}, O_{stop})$, which is calculated by softmax. The angle is represented by $a$ neurons $O_{angle} = (o_{1},...,o_{a})$ in range $[0, 2\pi]$, which is calculated by sigmoid. If $O_{walk}$ is larger than the pre-determined threshold $T$, the algorithm takes `walk' for the corresponding angle $\rm{arg max}_{i}(o_{i})$.

\section{RoadTracer using Adaptive DBN}
\label{sec:roadtracer_dbn}
In our previous research, the Adaptive DBN has been evaluated on the general classification and detection tasks. On the other hand, the RoadTracer is a different kind of task from them. In this paper, our motivation is to evaluate the performance of Adaptive DBN is superior to the original CNN for the road map detection.

\subsection{Training decision function}
\label{subsec:exe_train}
In this paper, the graph search algorithm of the RoadTracer was implemented on our Adaptive DBN instead of the CNN \cite{Bastani18} in order to improve the detection accuracy and the calculation time. Of course, the intermediate hidden layers are automatically constructed for input data by the self-organization function of our Adaptive DBN. The input and output layers of Adaptive DBN are implemented based on the specification of RoadTracer as mentioned in the section \ref{subsec:roadtracer_cnn}.

In the experiments, 25 cities of satellite images such as Chicago were used for the training process of the Adaptive DBN as same as the original paper \cite{Bastani18}. The satellite images and graph data in 25 cities were obtained from Google Map API and OpenStreetMap \cite{OpenStreetMap}, respectively. The area of each city was about 24 square km and the resolution of the satellite image was 60cm per 1 pixel. The size of input image to the model was $512 \times 512 \times 3$ RGB image. We used the following GPU workstation to the experiments: Nvidia GeForce RTX 3090 $\times 2$, Intel(R) Core(TM) i9-10900K CPU @ 3.70GHz, 64GB RAM, in Nvidia Docker.

For the training data, the detection accuracy was 94.7\% and 95.4\% for the CNN and the Adaptive DBN, respectively. The trained DBN was automatically formed with the network of 542, 502, 474, 298, 102, and 95 neurons from input to output layer. The calculation time was seven days for the CNN and five days for the Adaptive DBN, respectively. Although there was no significant difference between two models for the detection accuracy, the Adaptive DBN realized faster calculation time than the CNN. The trained models were evaluated on the test data in the section \ref{sec:exe_test}.

\subsection{Evaluation for Hiroshima}
\label{sec:exe_test}
\begin{figure}[tbp]
  \centering
  \includegraphics[scale=0.39]{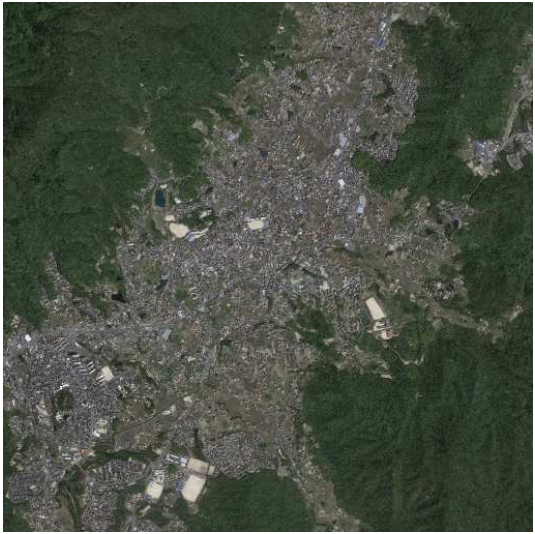}
  \vspace{-3mm}
  \caption{Test image (Kumano town)}
  \label{fig:kumano_org}
\vspace{-3mm}
\end{figure}

In this paper, we challenge to evaluate the performance of our model using a satellite image in the suburban area, since the original RoadTracer \cite{Bastani18} was evaluated using the satellite image of the urban area such as Chicago. Fig.~\ref{fig:kumano_org} shows the satellite image for test in this experiment, where is Kumano town, Hiroshima, Japan. The image was also obtained from Google Map API and the resolution was $4,096 \times 4,096$. The graph data for ground-truth was also obtained from OpenStreetMap. In the experiments, the precision and recall were used for evaluation method as follows.
\begin{equation}
\label{eq:precision}
Precision = \frac{TP}{TP+FP},
\end{equation}
\begin{equation}
\label{eq:recall}
Recall = \frac{TP}{TP+FN},
\end{equation}
where TP, FP, and FN are True Positive, False Positive, and False Negative, respectively. Precision means the ratio that the predicted vertices in road map are actually correct for the ground-truth ones. Recall means the ratio that the ground-truth vertices are actually detected by the model.

Table~\ref{tab:result} shows the detection accuracy and searching time for the test image. The performance of our Adaptive DBN was compared with the conventional CNN \cite{Bastani18}. The experiments were conducted with the different threshold $T = \{0.3, 0.1\}$. The stop action will be more decided in the searching algorithm if $T$ is increased.

From Table~\ref{tab:result}, our Adaptive DBN showed higher detection accuracy for not only Precision but also Recall than the CNN. In addition, the inference time was also lower than the CNN except $T = 0.1$. Although the searching time of the Adaptive DBN with $T = 0.3$ was twice as long as $T = 0.1$, the recall was improved with 5.9\%. We considered the searching area was wider by changing the parameter $T$ from 0.3 to 0.1. On the other hand, such improvement was not achieved in the CNN, regardless of the different parameter of $T$.

\begin{table}[tbp]
\caption{Detection Accuracy}
\vspace{-5mm}
\label{tab:result}
\begin{center}
\begin{tabular}{l|r|r|r}
\hline 
Model & Precision & Recall & Time(minutes) \\ \hline
CNN \cite{Bastani18} ($T = 0.3$)  & 74.4\% & 69.5\% & 49.2 \\
CNN \cite{Bastani18} ($T = 0.1$)  & 72.3\% & 70.3\% & 55.9 \\
Adaptive DBN ($T = 0.3$)  & 80.2\% & 85.8\% & {\bf 35.4} \\
Adaptive DBN ($T = 0.1$)  & {\bf 81.8\%} & {\bf 91.7\%} & 70.2 \\
\hline
\end{tabular}
\end{center}
\end{table}

\begin{figure}[tbp]
  \centering
  \subfigure[CNN ($T = 0.3$)]{\includegraphics[scale=0.35]{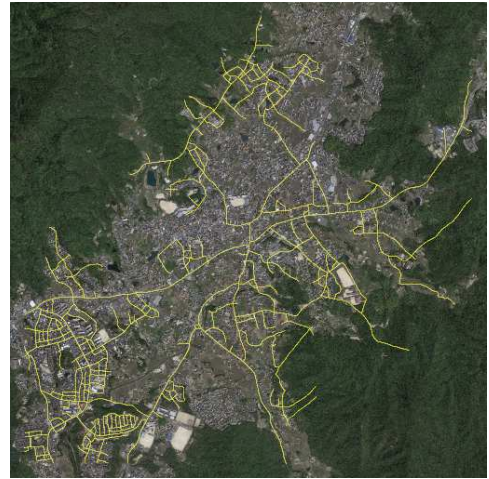}\label{fig:kumano_detected_cnn1}}
  \subfigure[CNN ($T = 0.1$)]{\includegraphics[scale=0.35]{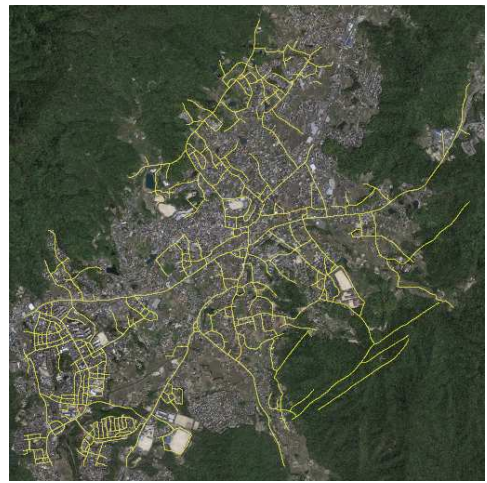}\label{fig:kumano_detected_cnn2}}
  \vspace{-3mm}
  \caption{Detection results of CNN}
  \label{fig:kumano_result_cnn}
\end{figure}

\begin{figure}[tbp]
  \centering
  \subfigure[Adaptive DBN ($T = 0.3$)]{\includegraphics[scale=0.35]{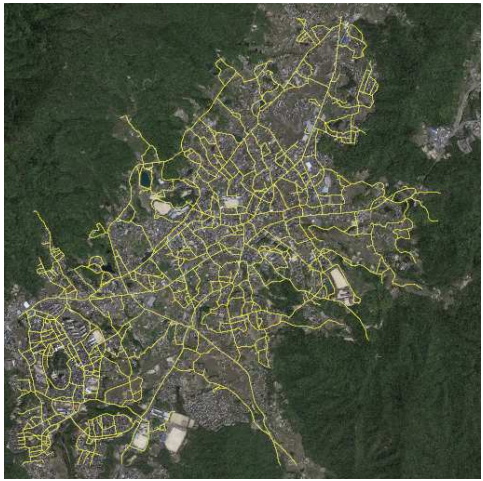}\label{fig:kumano_detected_dbn1}}
  \subfigure[Adaptive DBN ($T = 0.1$)]{\includegraphics[scale=0.35]{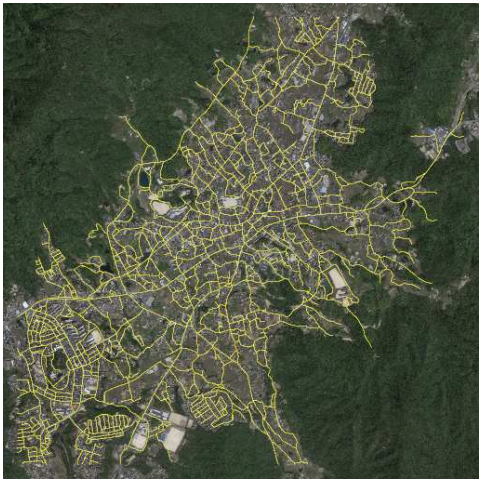}\label{fig:kumano_detected_dbn2}}
  \vspace{-3mm}
  \caption{Detection results of Adaptive DBN}
  \label{fig:kumano_result_dbn}
\vspace{-3mm}
\end{figure}

Fig.~\ref{fig:kumano_result_cnn} and Fig.~\ref{fig:kumano_result_dbn} show the detection results for Kumano town by the CNN and the Adaptive DBN, respectively. The yellow lines on the satellite image is the detected road map. We can see the Adaptive DBN was able to detect more roads than the CNN.

\begin{figure}[tbp]
  \centering
  \subfigure[Case1]{\includegraphics[scale=0.4]{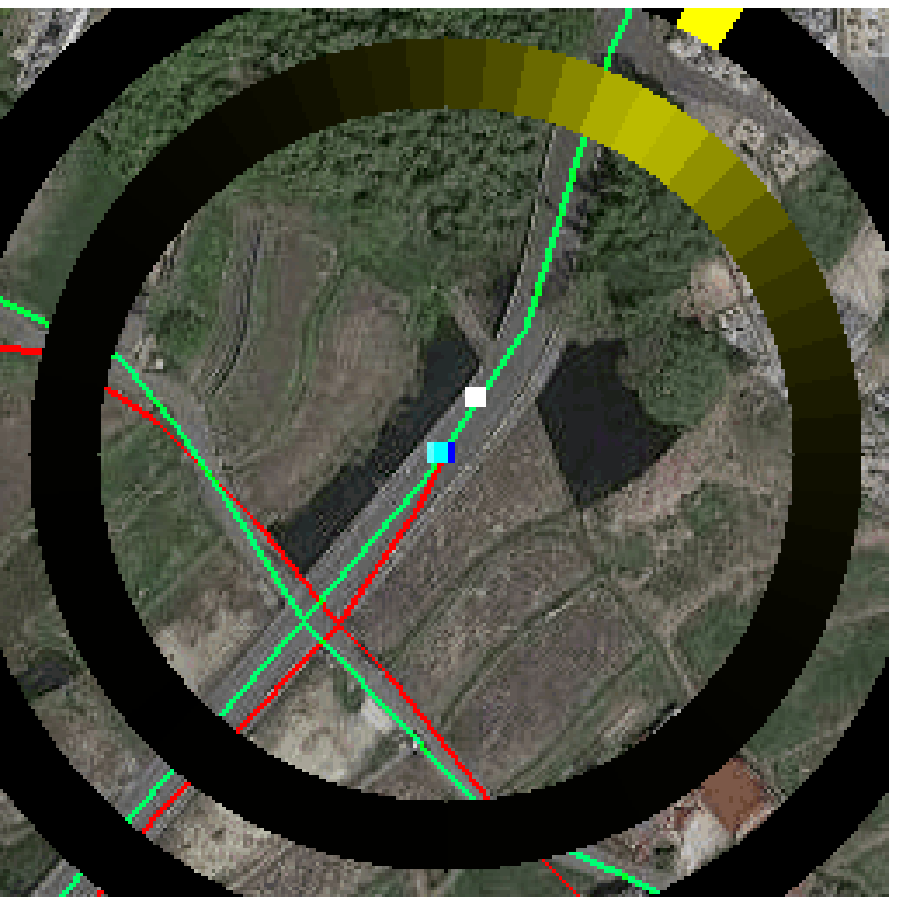}\label{fig:case_tp_1}}
  \subfigure[Case2]{\includegraphics[scale=0.4]{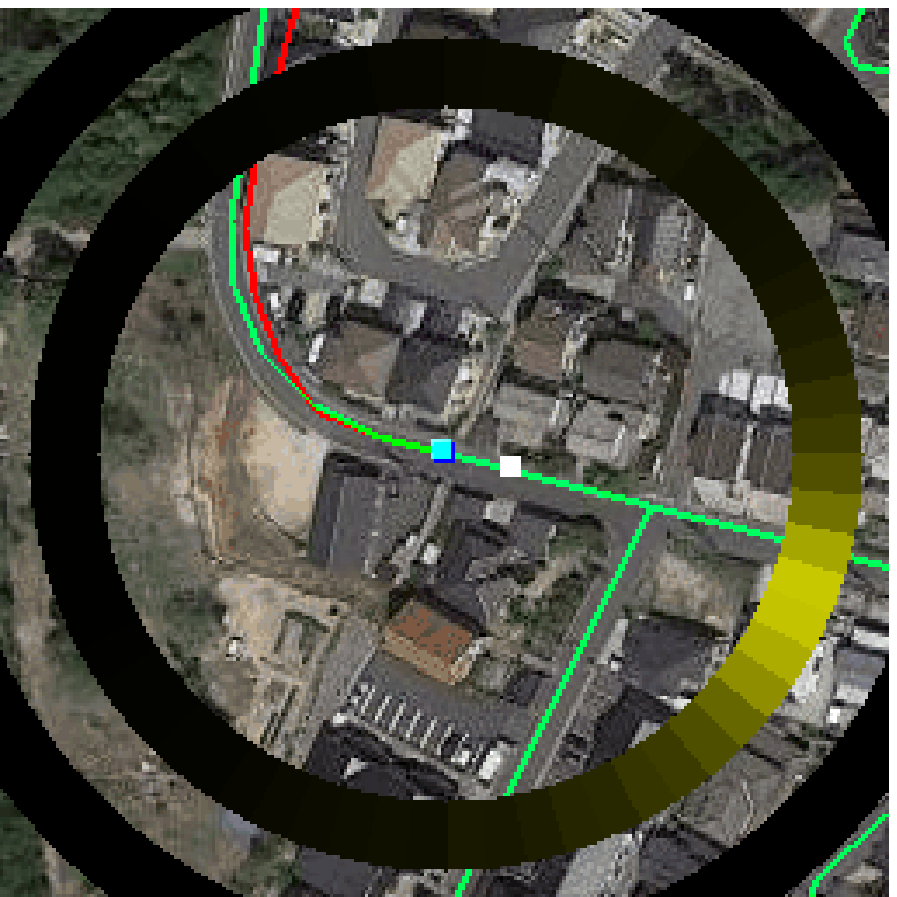}\label{fig:case_tp_2}}
  \subfigure[Case3]{\includegraphics[scale=0.4]{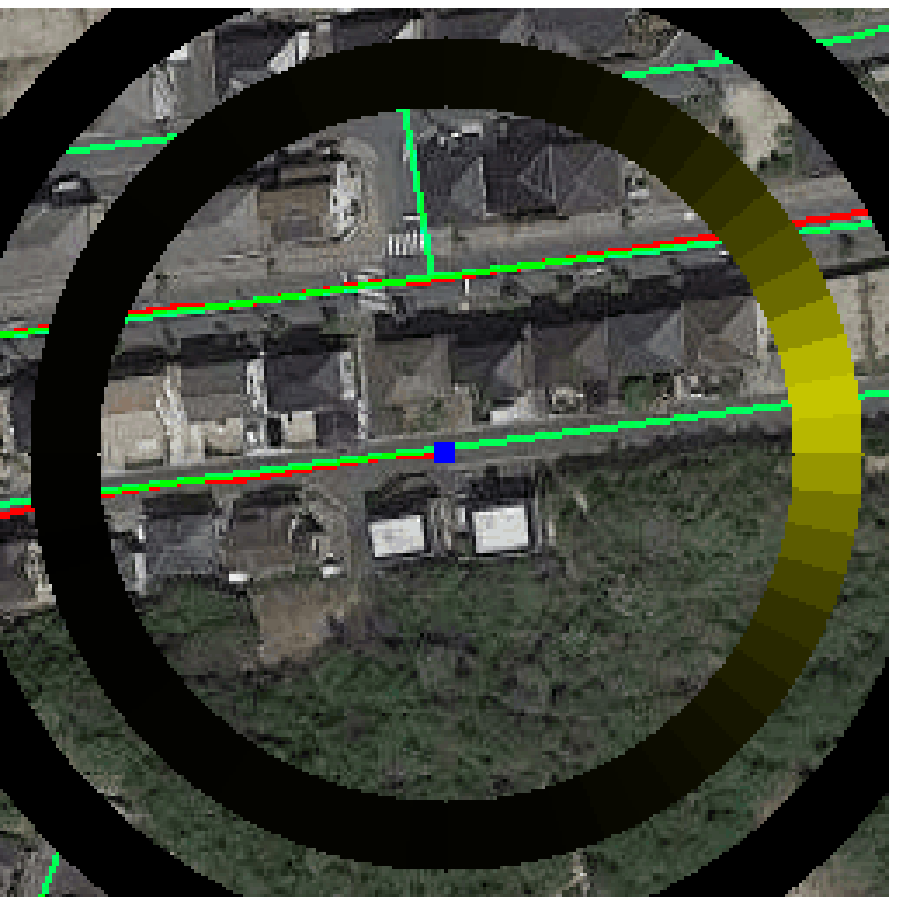}\label{fig:case_tp_3}}
  \subfigure[Case4]{\includegraphics[scale=0.4]{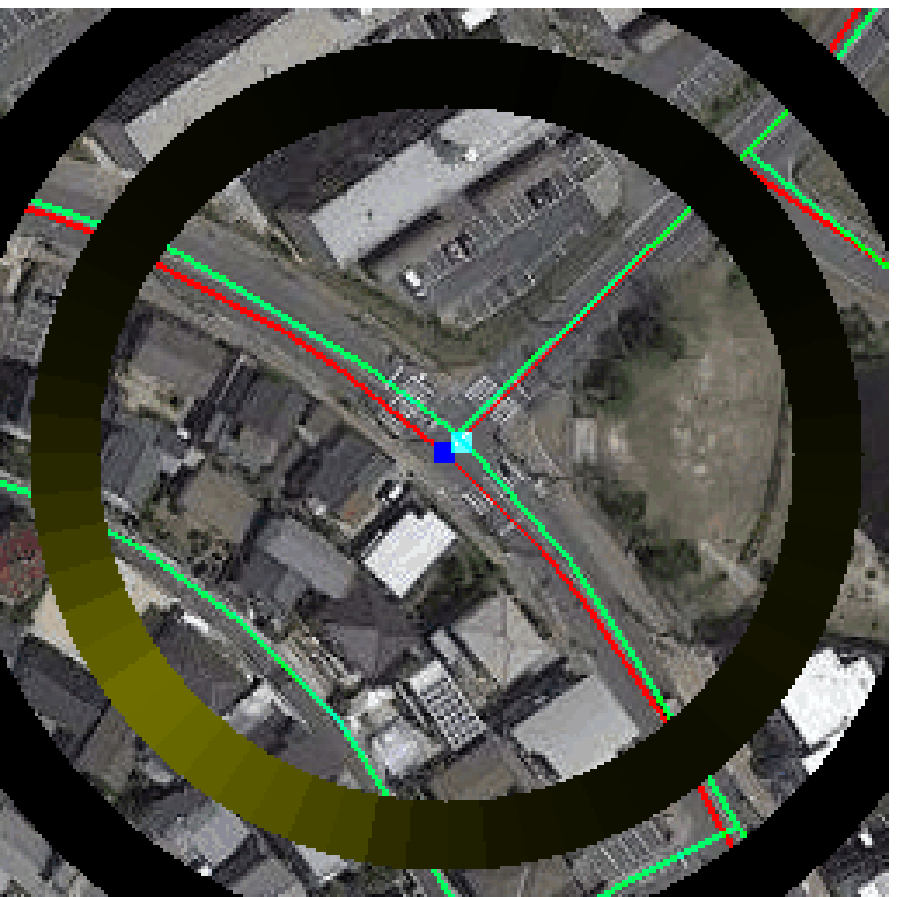}\label{fig:case_tp_4}}
  \vspace{-3mm}
  \caption{True Positive cases by Adaptive DBN}
  \label{fig:kuman_tp}
\vspace{-3mm}
\end{figure}

\begin{figure*}[tbp]
  \centering
  \subfigure[Case 1-1]{\includegraphics[scale=0.4]{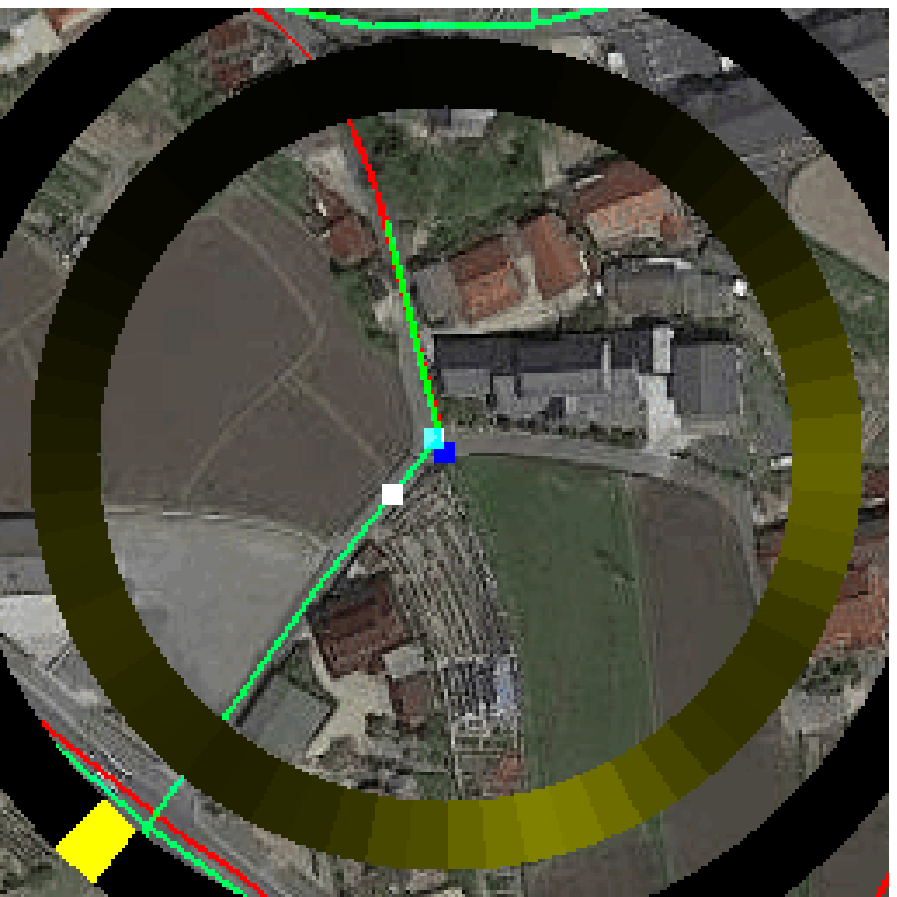}\label{fig:case_fp_1_1}}
  \subfigure[Case 1-2]{\includegraphics[scale=0.4]{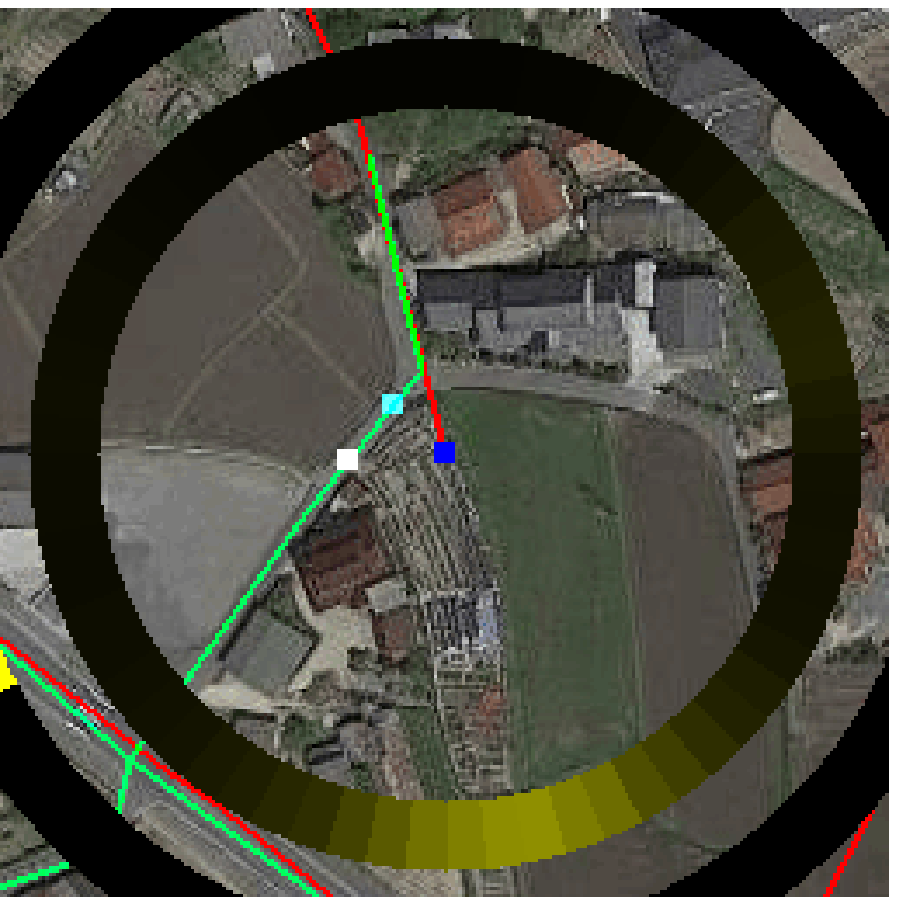}\label{fig:case_fp_1_2}}
  \subfigure[Case 1-3]{\includegraphics[scale=0.4]{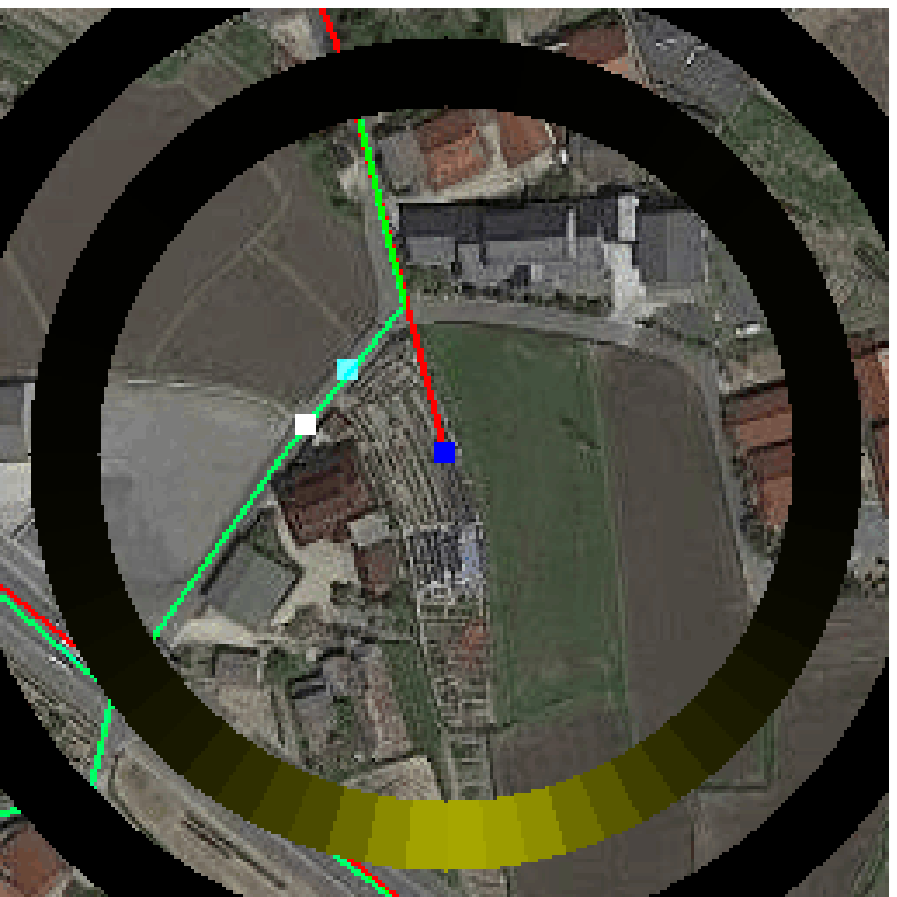}\label{fig:case_fp_1_3}}
  \subfigure[Case 1-4]{\includegraphics[scale=0.4]{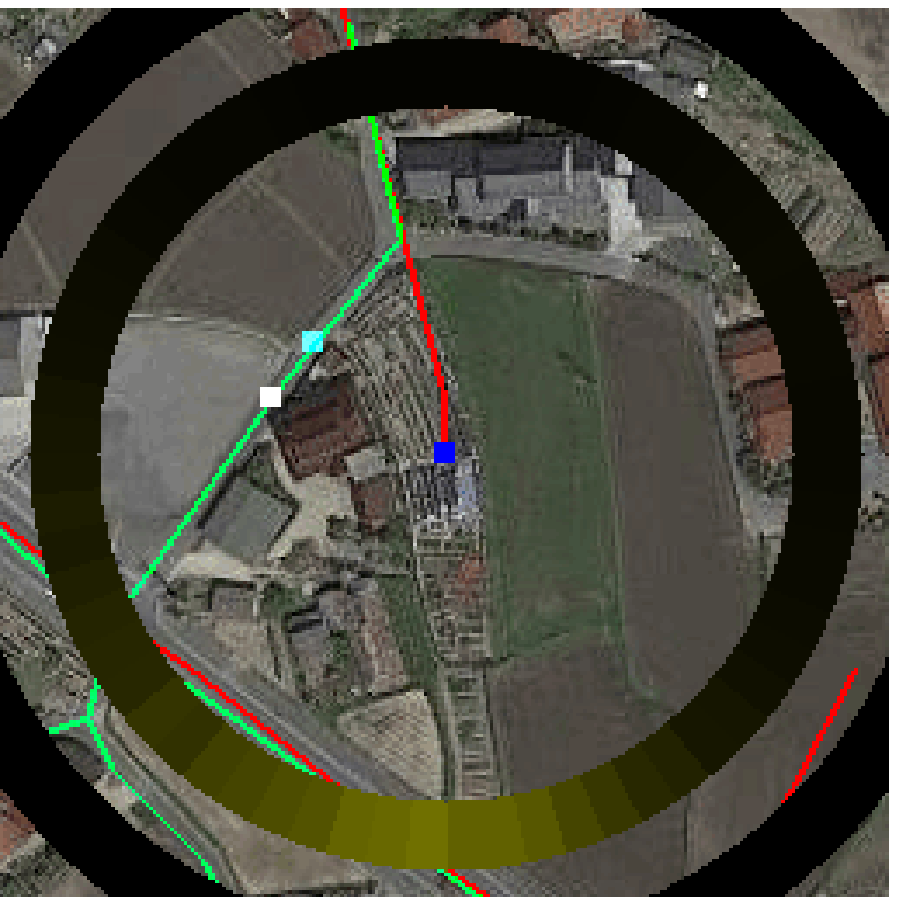}\label{fig:case_fp_1_4}}
  \subfigure[Case 2-1]{\includegraphics[scale=0.4]{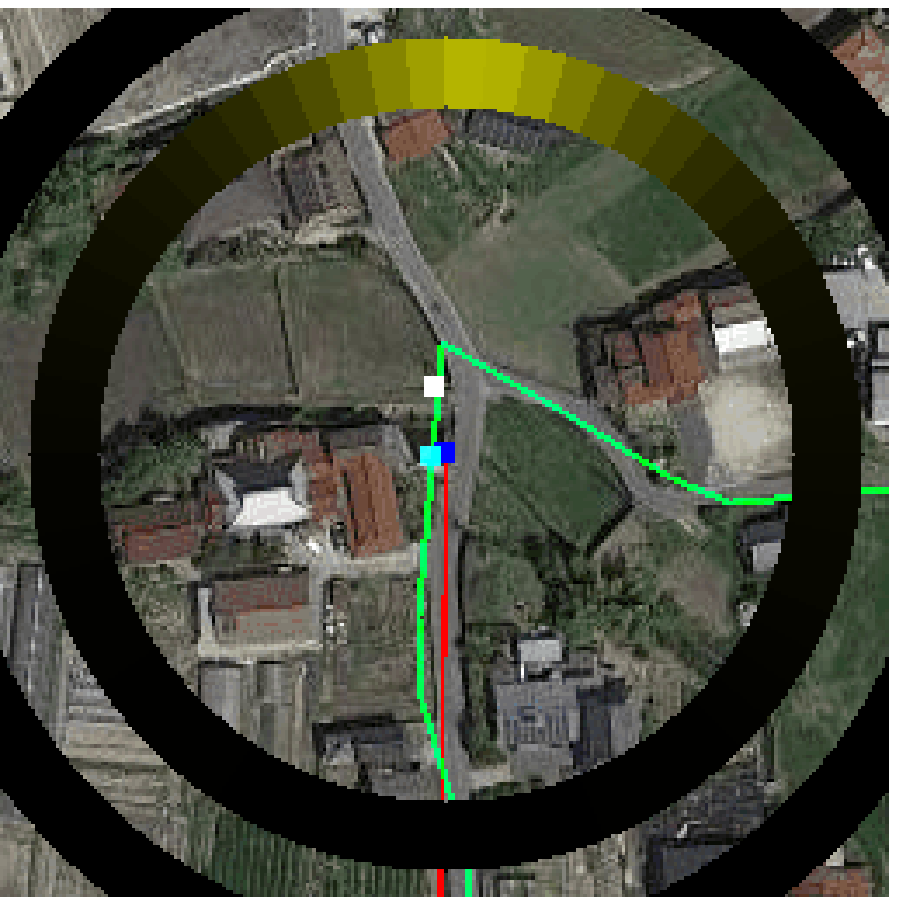}\label{fig:case_fp_2_1}}
  \subfigure[Case 2-2]{\includegraphics[scale=0.4]{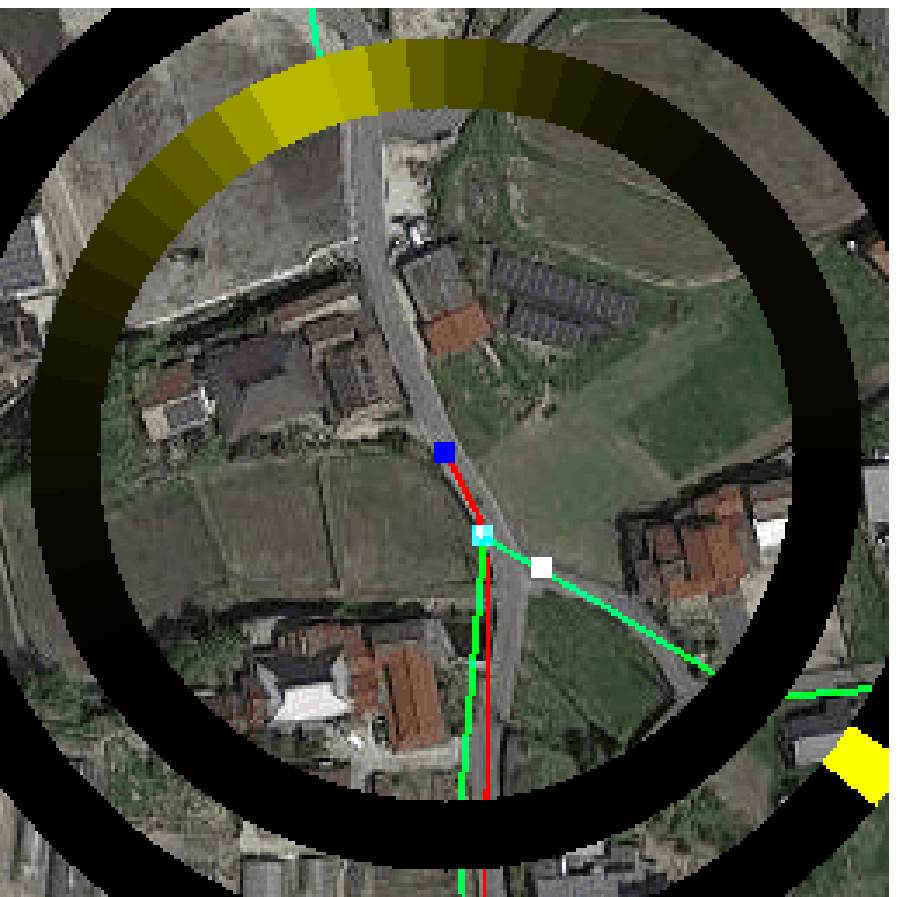}\label{fig:case_fp_2_2}}
  \subfigure[Case 2-3]{\includegraphics[scale=0.4]{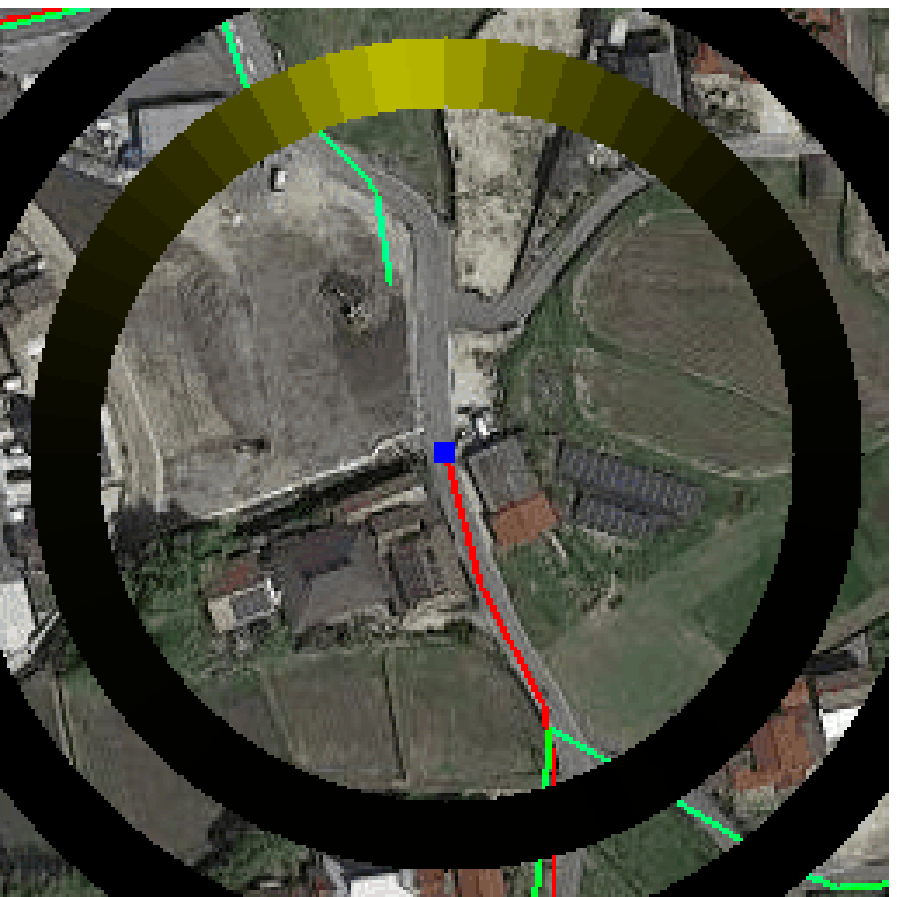}\label{fig:case_fp_2_3}}
  \subfigure[Case 2-4]{\includegraphics[scale=0.4]{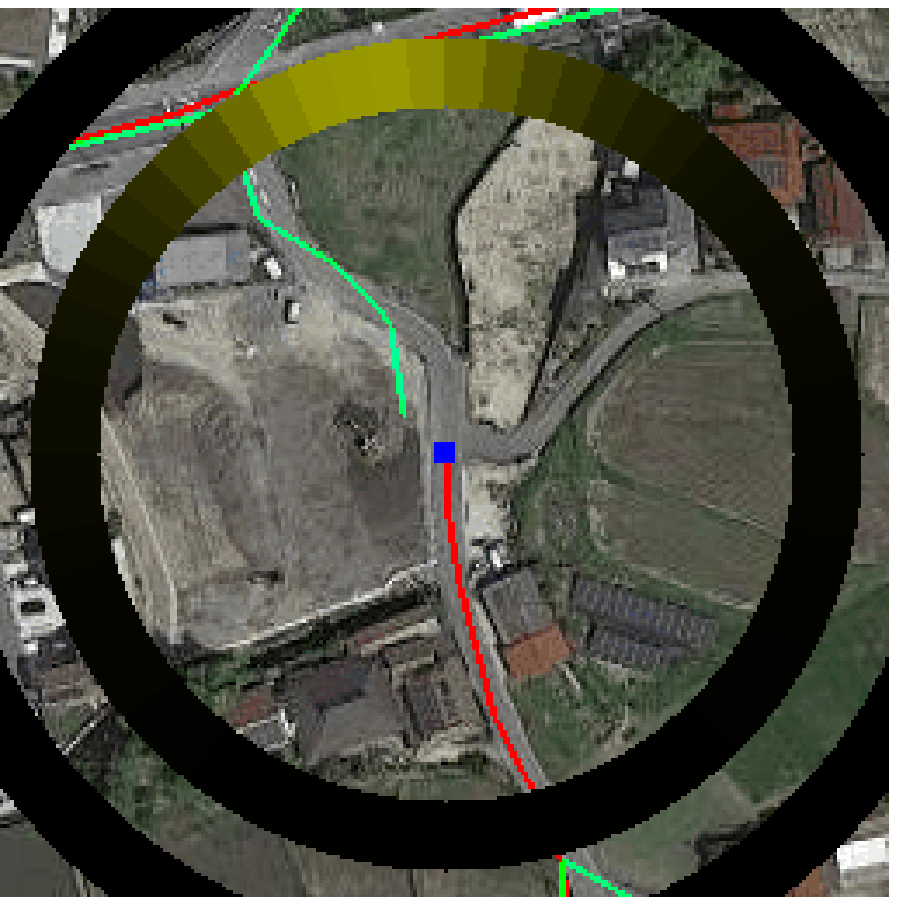}\label{fig:case_fp_2_4}}
  \subfigure[Case 3-1]{\includegraphics[scale=0.4]{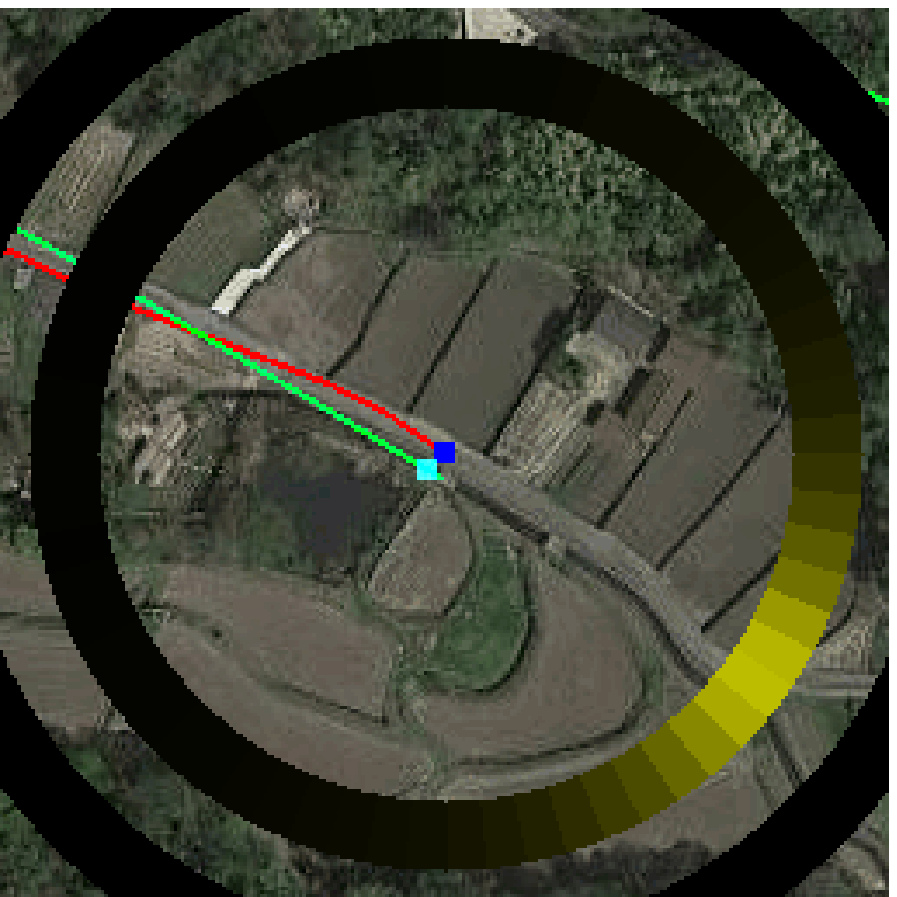}\label{fig:case_fp_3_1}}
  \subfigure[Case 3-2]{\includegraphics[scale=0.4]{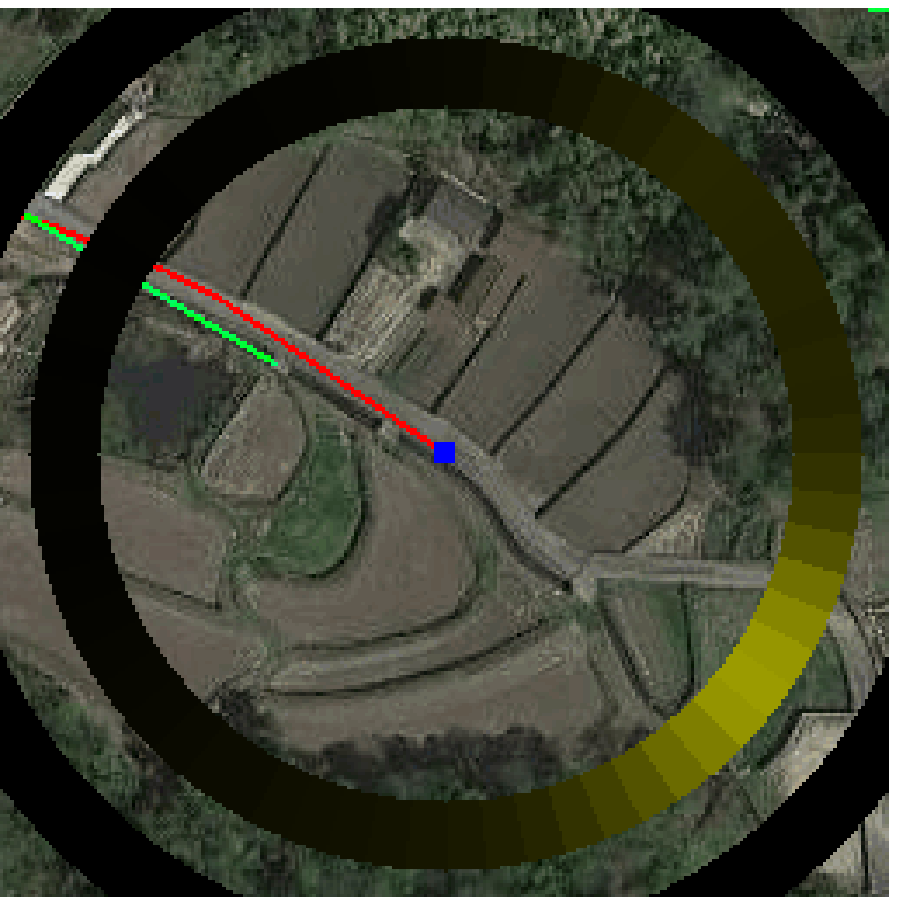}\label{fig:case_fp_3_2}}
  \subfigure[Case 3-3]{\includegraphics[scale=0.4]{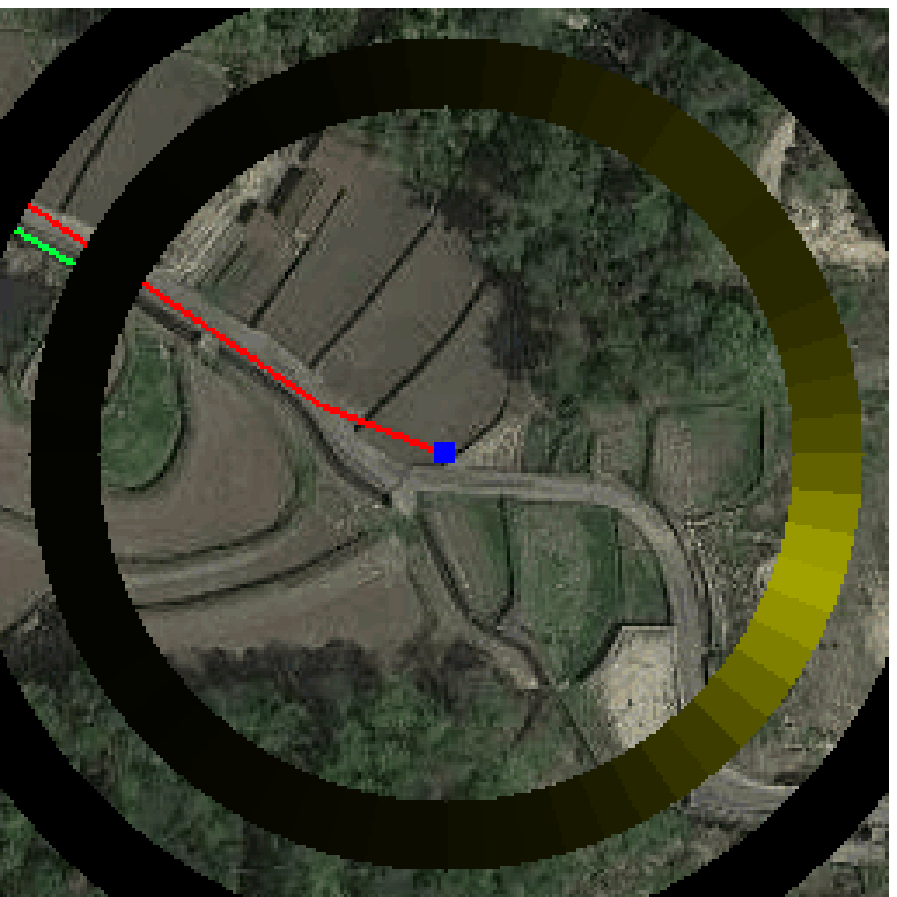}\label{fig:case_fp_3_3}}
  \subfigure[Case 3-4]{\includegraphics[scale=0.4]{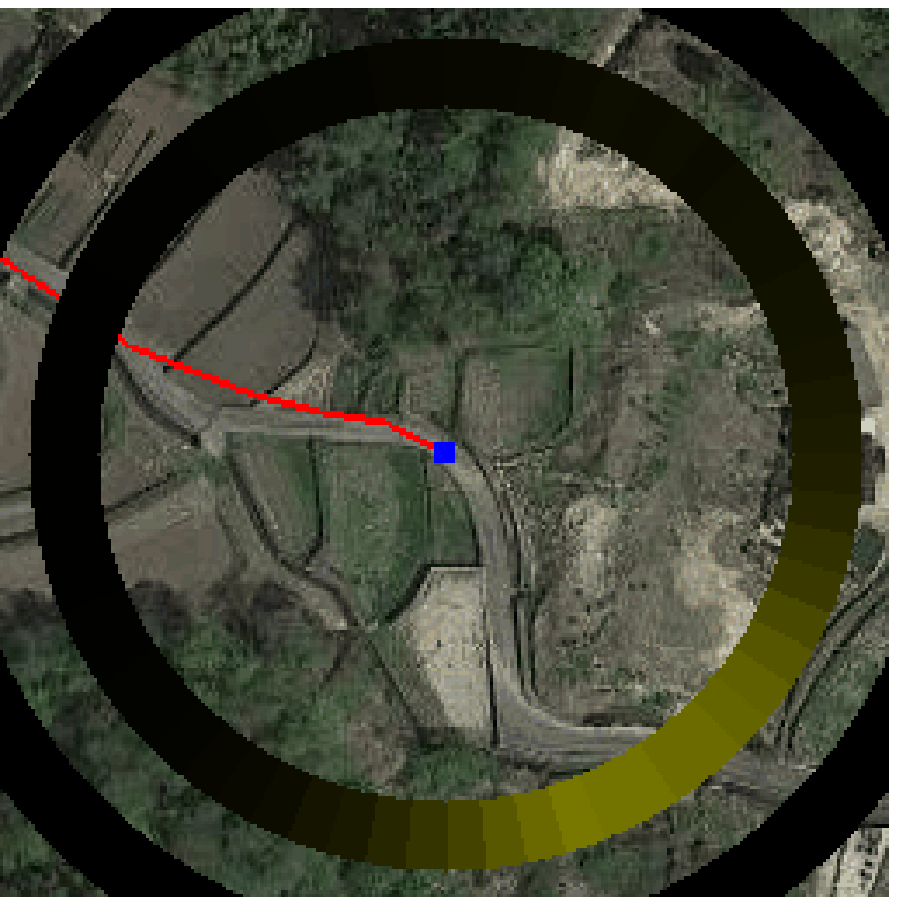}\label{fig:case_fp_3_4}}
  \subfigure[Case 4-1]{\includegraphics[scale=0.4]{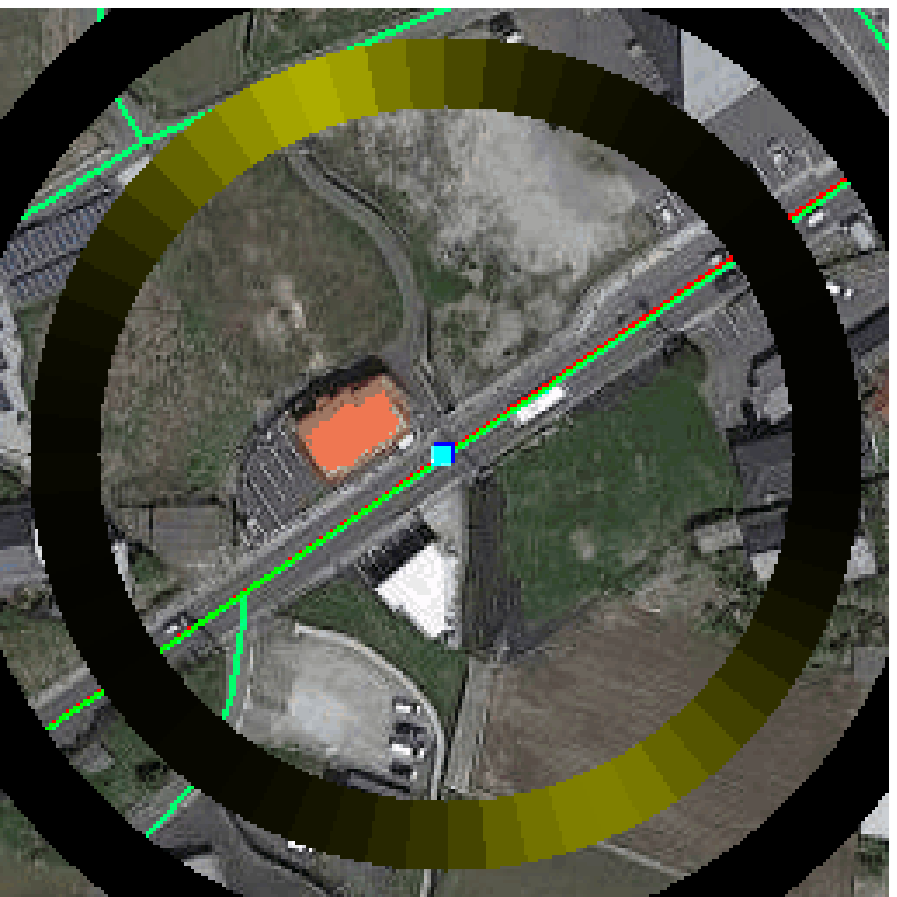}\label{fig:case_fp_4_1}}
  \subfigure[Case 4-2]{\includegraphics[scale=0.4]{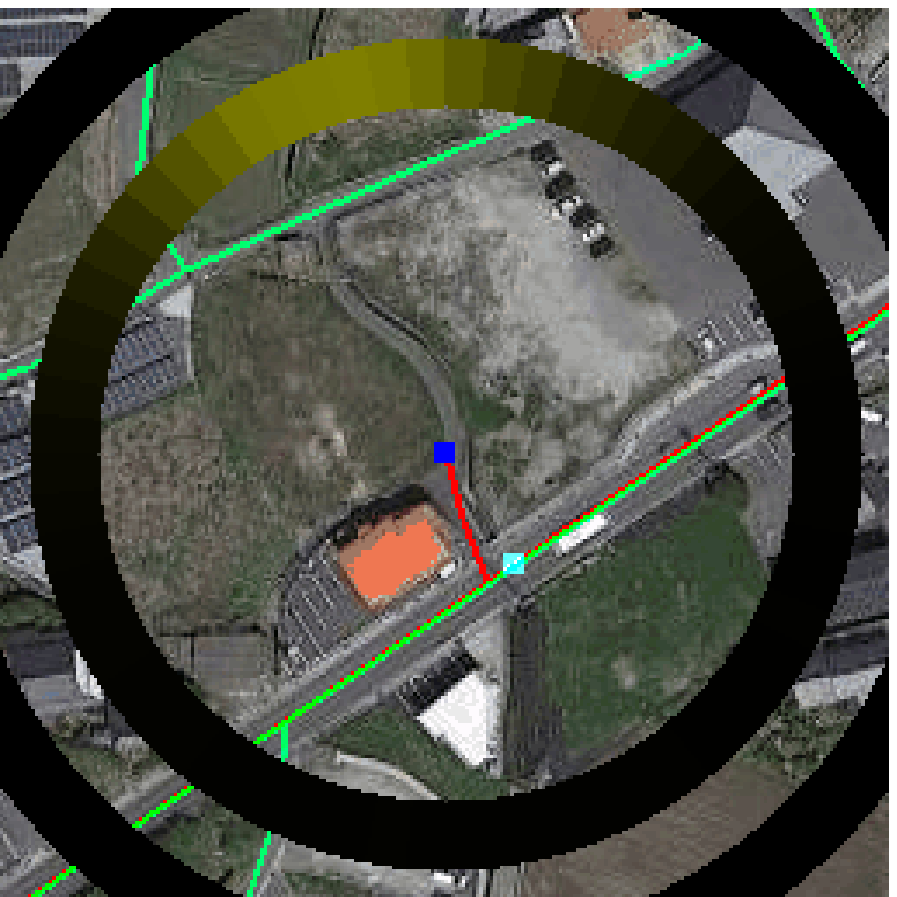}\label{fig:case_fp_4_2}}
  \subfigure[Case 4-3]{\includegraphics[scale=0.4]{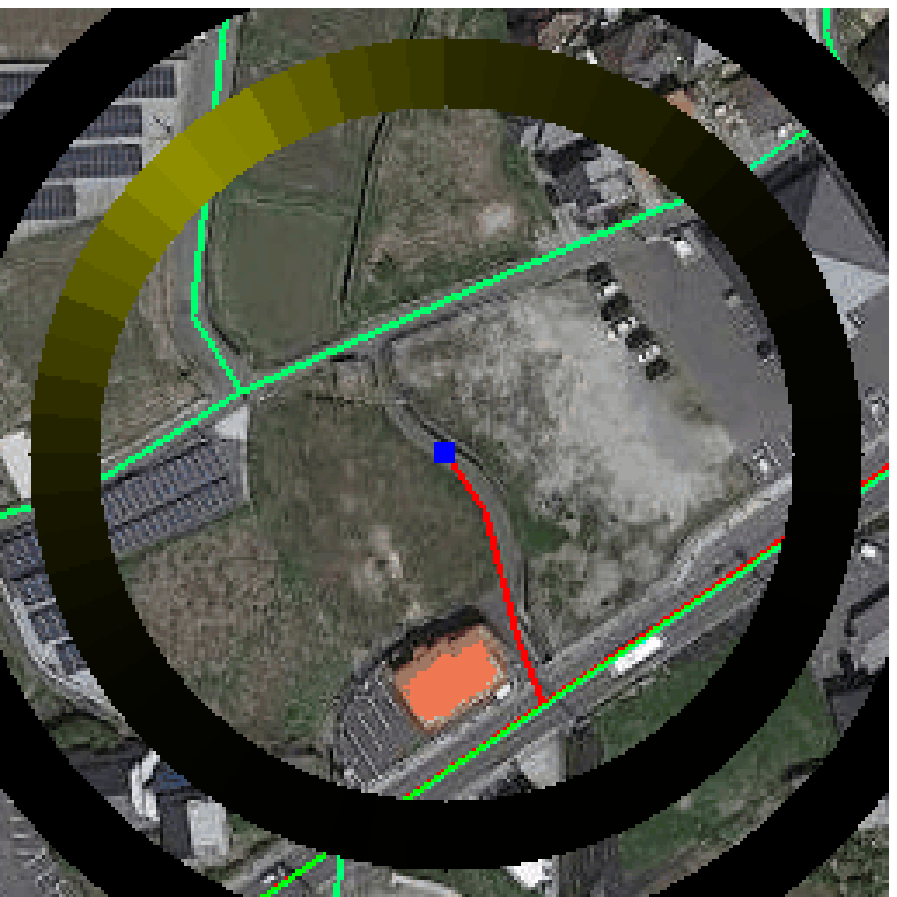}\label{fig:case_fp_4_3}}
  \subfigure[Case 4-4]{\includegraphics[scale=0.4]{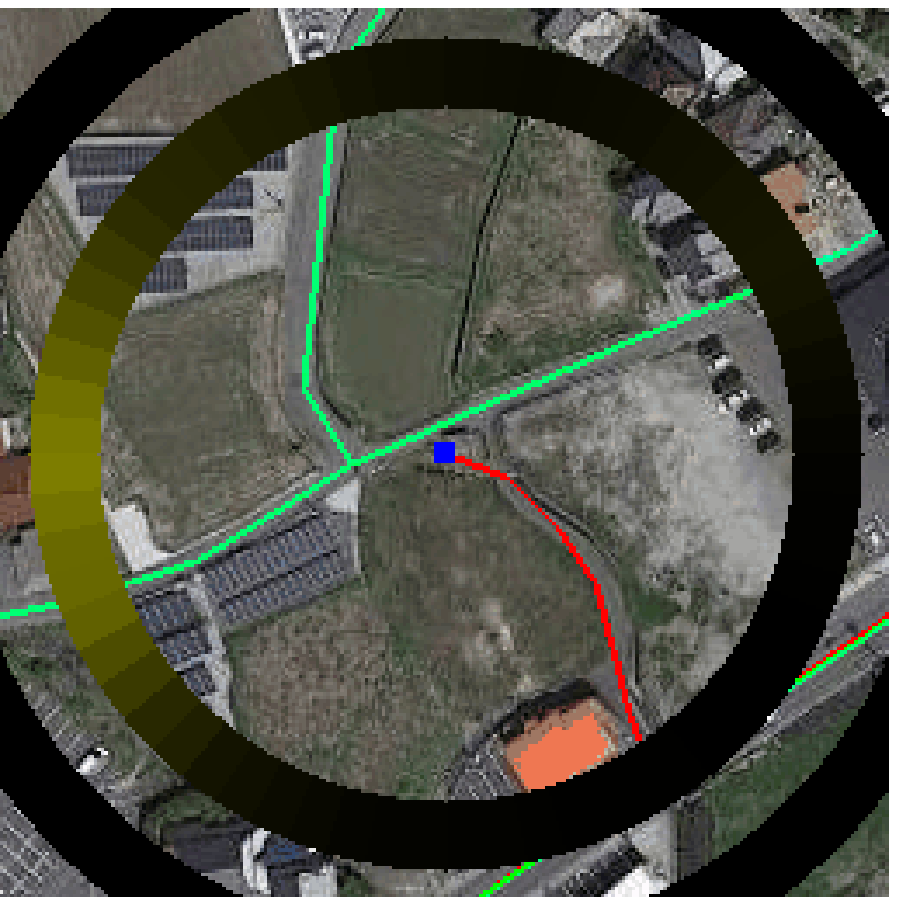}\label{fig:case_fp_4_4}}
  \subfigure[Case 5-1]{\includegraphics[scale=0.4]{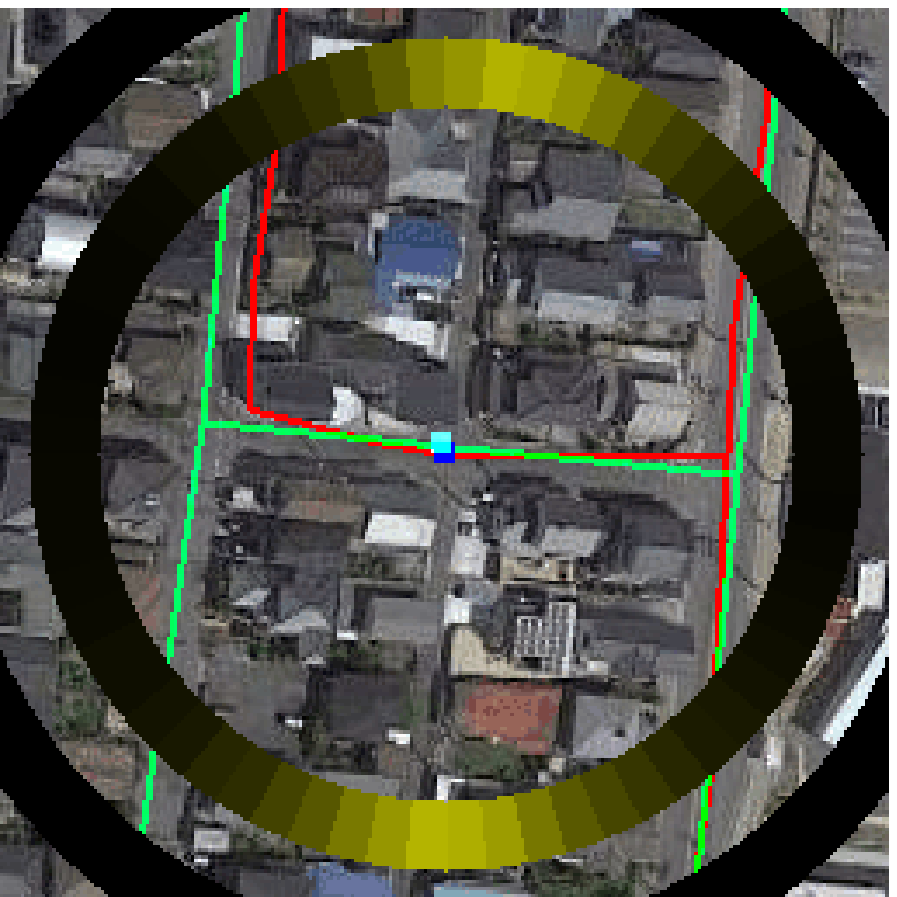}\label{fig:case_fp_5_1}}
  \subfigure[Case 5-2]{\includegraphics[scale=0.4]{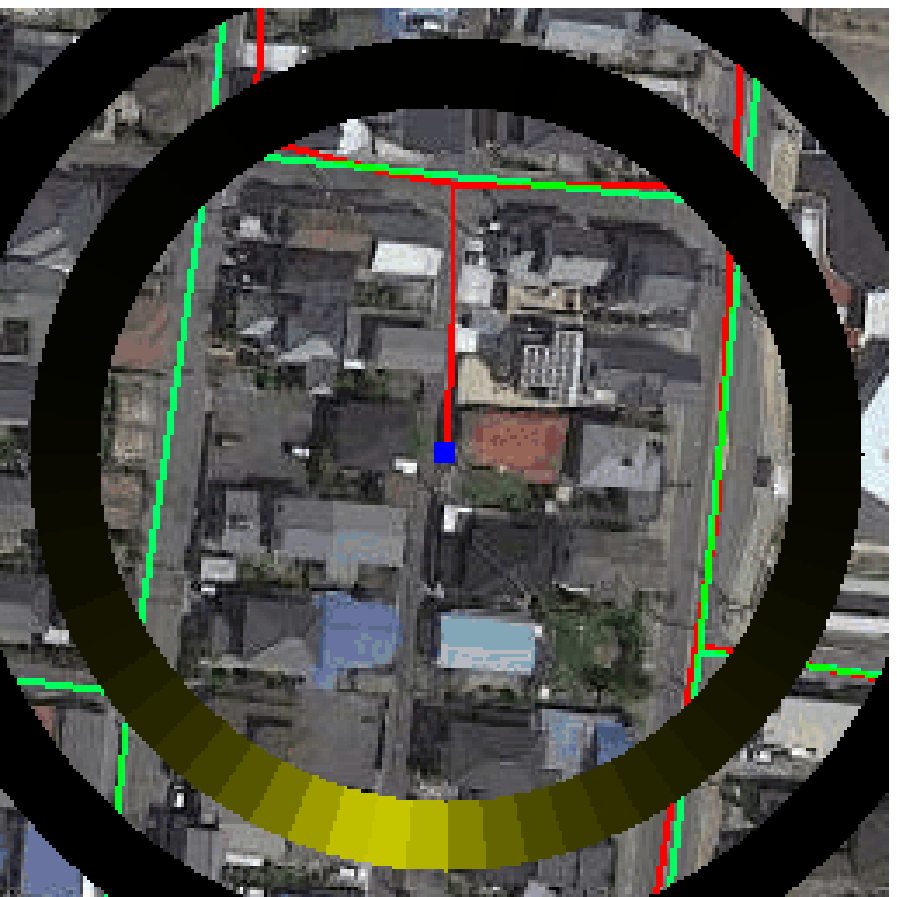}\label{fig:case_fp_5_2}}
  \subfigure[Case 5-3]{\includegraphics[scale=0.4]{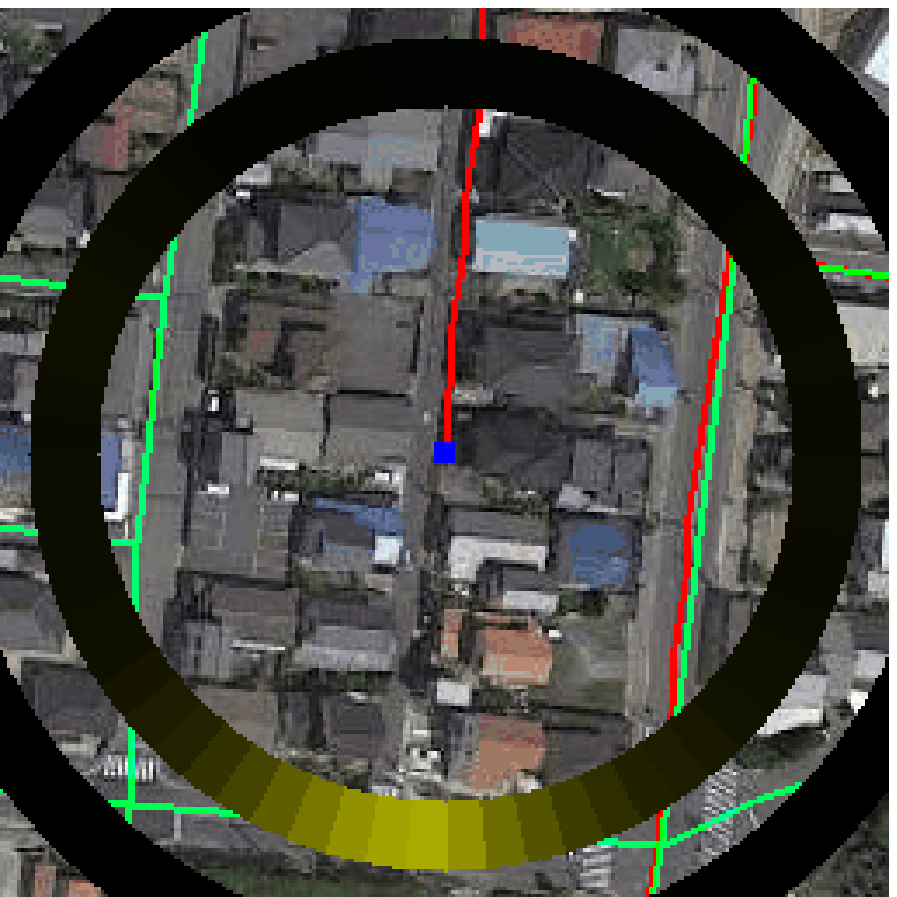}\label{fig:case_fp_5_3}}
  \subfigure[Case 5-4]{\includegraphics[scale=0.4]{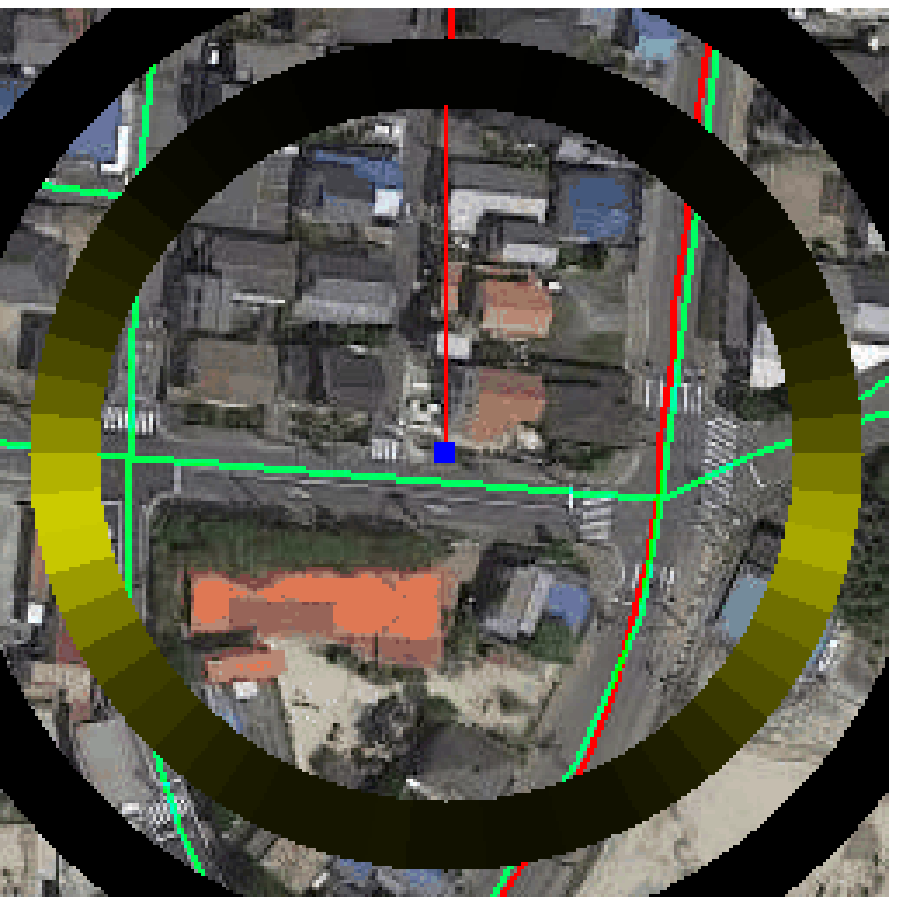}\label{fig:case_fp_5_4}}  
  \caption{False Positive cases by Adaptive DBN}
  \label{fig:kuman_fp}
\end{figure*}

By changing the threshold $T$ from 0.3 to 0.1, the walk action was more likely decided in the searching algorithm. This is, if the threshold $T$ is small, the searching algorithm is worked to make a deep exploration around the intersection, although the infrence time was varying very widely. As a result, the algorithm was able to detect more True Positive cases, while more False Positive cases were also found. However, we noted some False Positive cases might be actually road features. Fig.~\ref{fig:kuman_tp} shows the True Positive cases by the Adaptive DBN with $T = 0.1$. The green and red lines in the figures are the ground truth and the detected road maps, respectively. The two filled circles represent the search direction with black to yellow gradation, where outside and inside ones are ground-truth and the predicted ones, respectively. Fig.~\ref{fig:kuman_fp} shows the same result related to the False Positive cases. The figures show five cases by four continuous searching result. As shown in Fig.~\ref{fig:kuman_tp}, the detected lines and the ground truth ones were almost matched in the True Positive cases. Although they were not matched in the False Positive cases, the detected lines might be actual roads as shown in Fig.~\ref{fig:kuman_fp}. For example, Fig.~\ref{fig:case_fp_2_1} to Fig~\ref{fig:case_fp_2_4} show a case that the detected roads might be actually roads, but not defined in OpenStreetMap. This is, Adaptive DBN was able to detect narrow roads such as farm roads and instructional roads that were not defined in OpenStreetMap. We will investigate such mis-detected results by the other additional data and expert knowledge.


\section{Conclusion}
\label{sec:conclusion}
Adaptive DBN is an adaptive structure learning method of DBN which can find an optimal network structure by generating / annihilating neurons and hierarchization during learning. In this paper, our model is applied to an automatic recognition method of road network, RoadTracer. RoadTracer can generate a road map on the ground surface from aerial photograph data. In the search algorithm of network graph, a CNN is trained to find network connectivity between roads with high detection capability. However, the system takes a long calculation time for not only the training phase but also the inference phase, then it may not reach high accuracy. In order to improve the accuracy and the calculation time, our Adaptive DBN was implemented on the RoadTracer instead of the CNN. The performance of our developed model was evaluated on a satellite image in the suburban area, Japan. Our Adaptive DBN had an advantage of not only the detection accuracy but also the inference time compared with the conventional CNN in the experiment results. To improve the detection accuracy of our model, the detailed detection results using the other cities will be investigated in future.

\section*{Acknowledgment}
This work was supported by JSPS KAKENHI Grant Number 19K12142, 19K24365, 21K17809, and National Institute of Information and Communications Technology (NICT), JAPAN.


\begin{thebibliography}{1}

\bibitem{Bengio09}
Y.Bengio (2009) \emph{Learning Deep Architectures for AI}, Foundations and Trends in Machine Learning archive, vol.2, no.1, pp.1--127
  
\bibitem{Kamada18_Springer}
S.Kamada, T.Ichimura, A.Hara, and K.J.Mackin, \emph{Adaptive Structure Learning Method of Deep Belief Network using Neuron Generation-Annihilation and Layer Generation}, Neural Computing and Applications, pp.1--15 (2018)

\bibitem{Hinton06}
G.E.Hinton, S.Osindero and Y.Teh, \emph{A fast learning algorithm for deep belief nets}, Neural Computation, vol.18, no.7, pp.1527--1554 (2006)
  
\bibitem{Hinton12}
 G.E.Hinton, \emph{A Practical Guide to Training Restricted Boltzmann Machines}, Neural Networks, Tricks of the Trade, Lecture Notes in Computer Science (LNCS, vol.7700), pp.599--619 (2012)

\bibitem{Kamada16_SMC}
S.Kamada and T.Ichimura, \emph{An Adaptive Learning Method of Restricted Boltzmann Machine by Neuron Generation and Annihilation Algorithm}. Proc. of 2016 IEEE International Conference on Systems, Man, and Cybernetics (SMC2016), pp.1273--1278 (2016)  
  
\bibitem{Kamada16_ICONIP}
S.Kamada, T.Ichimura, \emph{A Structural Learning Method of Restricted Boltzmann Machine by Neuron Generation and Annihilation Algorithm}, Neural Information Processing, Proc. of the 23rd International Conference on Neural Information Processing, Springer LNCS9950), pp.372--380 (2016)

\bibitem{Kamada16_TENCON}
S.Kamada and T.Ichimura, \emph{An Adaptive Learning Method of Deep Belief Network by Layer Generation Algorithm}, Proc. of IEEE TENCON2016, pp.2971--2974 (2016)
 
\bibitem{CIFAR10}
A.Krizhevsky, \emph{Learning Multiple Layers of Features from Tiny Images}, Master of thesis, University of Toronto (2009)

\bibitem{AlexNet}
A.Krizhevsky, I.Sutskever, G.E.Hinton, \emph{ImageNet Classification with Deep Convolutional Neural Networks}, Proc. of Advances in Neural Information Processing Systems 25 (NIPS 2012) (2012)

\bibitem{VGG16}
K.Simonyan, A.Zisserman, \emph{Very deep convolutional networks for large-scale image recognition}, Proc. of International Conference on Learning Representations (ICLR 2015) (2015)
  
\bibitem{GoogLeNet}
C.Szegedy, W. Liu, Y.Jia, P.Sermanet, S.Reed, D.Anguelov, D.Erhan, V.Vanhoucke, A.Rabinovich, \emph{Going Deeper with Convolutions}, Proc. of CVPR2015 (2015)

\bibitem{ResNet}
K.He, X.Zhang, S.Ren, J.Sun, J, \emph{Deep residual learning for image recognition}, Proc. of 2016 IEEE Conference on Computer Vision and Pattern Recognition (CVPR), pp.770--778 (2016)

\bibitem{Liu18}
Y.Liu, et.al., \emph{RoadNet: Learning to Comprehensively Analyze Road Networks in Complex Urban Scenes From High-Resolution Remotely Sensed Images}, IEEE Transactions on Geoscience and Remote Sensing, vol.57, no.4, pp.2043--2056 (2018)

\bibitem{He20}
S.He, et.al., \emph{RoadTagger: Robust Road Attribute Inference with Graph Neural Networks}, Proc. of the AAAI Conference on Artificial Intelligence, vol.34, no.07, pp.10965-10972 (2020)

\bibitem{Lian20}
R.Lian and L.Huang, \emph{DeepWindow: Sliding Window Based on Deep Learning for Road Extraction From Remote Sensing Images}, IEEE Journal of Selected Topics in Applied Earth Observations and Remote Sensing, vol.13, pp.1905-1916 (2020)

\bibitem{Tan20}
Y.Q.Tan, et.al., \emph{VecRoad: Point-Based Iterative Graph Exploration for Road Graphs Extraction}, Proc. of 2020 IEEE/CVF Conference on Computer Vision and Pattern Recognition (CVPR), pp.8907-8915 (2020)

\bibitem{Mattyus17}
G.Máttyus, et.al., \emph{DeepRoadMapper: Extracting road topology from aerial images}, Proc. of the IEEE Conference on Computer Vision and Pattern Recognition (ICCV), pp.3438--3446 (2017)

\bibitem{Bastani18}
F.Bastani, et.al., \emph{RoadTracer: Automatic Extraction of Road Networks from Aerial Images}, arXiv:1802.03680 [cs.CV] (2018)

\bibitem{OpenStreetMap}
S.Coast, OpenStreetMap, \url{https://www.openstreetmap.org/} [online, 2021/3/15]

      
\end{thebibliography}
\end{document}